# A Statistical Method for Object Counting

Jans Glagoļevs and Kārlis Freivalds
Institute of Mathematics and Computer Science
University of Latvia
Riga, Latvia
{jans.glagolevs, karlis.freivalds}@lumii.lv

*Abstract*— **In this paper we present a new object counting method that is intended for counting similarly sized and mostly round objects. Unlike many other algorithms of the same purpose, the proposed method does not rely on identifying every object, it uses statistical data obtained from the image instead. The method is evaluated on images with human bone cells, oranges and pills achieving good accuracy. Its strengths are ability to deal with touching and partly overlapping objects, ability to work with different kinds of objects without prior configuration and good performance.**

## I. Background

Object counting in images and video streams is used in many fields including live cell counting from microscopic images and product counting on production lines. In this paper we present a new object counting algorithm that is intended for counting similarly sized and mostly round objects. Unlike many other algorithms of the same purpose, the proposed algorithm does not rely on identifying every object. It approximates the number of objects from the distribution characteristics of image pixels instead.

There exist several algorithms for counting objects with similar sizes and round shapes. It is common to use some segmentation method as the first step to produce a binary image consisting of white objects on a black background. Then, most of the known algorithms identify every object and count them. Watershed segmentation is one of the known methods for the given task [11]. It allows to calculate the number of convex objects; objects can overlap with each other and be differently sized. Watershed segmentation can be implemented by iteratively eroding an image (deleting boundary points of the possible objects) [6]. During erosion process touching convex objects separate and later disappear. The number of disappearances is used to estimate the number of objects. Slayton M. Costa and Suann Yang have developed a method that uses watershed segmentation from ImageJ (NIH) library to count pollen grains from microscopic images [2]. Another approach is to use a Circular Hough transform [3]. Pixels belonging to object boundaries are determined and a Hough transform is used to extract circles that cover a significant amount of the boundary pixels. A Circular Hough transform is also used by Monika Mogra, Arum Bansel and Vivek Srivastava to identify and count white blood cells during blood analysis [8].

Xiaomin Guo and Feihong Yu have a method for counting live cells that simultaneously uses dual threshold, watershed segmentation, blob analysis and k-means method [4]. Authors claim that the algorithm works well with different kinds of cells. Round object counting can also be done by using artificial neural networks (ANN). For example, Watcharin et al. use ANN for image segmentation; then they find objects with Hough transform [12]. Blood cell segmentation and counting can also be done by using Pulse-Coupled Neural Networks as described by Su Mao-jun et al. [7]. Navin D. Jambhekar describes red and white blood cell recognition in low resolution images by using only an ANN [5].

In this paper we present a new algorithm that uses a new approach for object counting. At first, an estimate of the size of the object is made based on the distribution of intervals belonging to objects in a chosen direction. Then, object count is calculated from the total area of the objects and their estimated size. We show that such approach yields a highly accurate object count estimate in several scenarios including cases with touching or overlapping objects where distinguishing individual objects is problematic.

## II. Object Size Estimation

Consider an image consisting of mostly round and similarly sized objects. We will develop several models of such images and see how to obtain the average size of objects in them. We will derive a counting algorithm based on these models. Let us start with a case when we have $n$ identical non-touching white discs on a black plane and we wish to identify their diameter. Such case can be solved trivially by obtaining the longest white interval, say, in horizontal direction and taking its length as the required diameter.

What happens if the discs are touching and possibly overlapping? Let us inspect the distribution of lengths of the intervals in the non-overlapping case, first. If we ignore rounding errors producing by discrete pixel grid, it is easy to see that the distribution function of the segment lengths $h_d(x)$ is proportional to the derivative of the circle equation, namely:

$$h_d(x) = \frac{x}{\sqrt{d^2 - x^2}}, \tag{1}$$

if $0 \leq x < d$, and 0 otherwise, where $d$ is the diameter of the circle. For a disc of diameter 60 the distribution function is shown in Fig. 1.

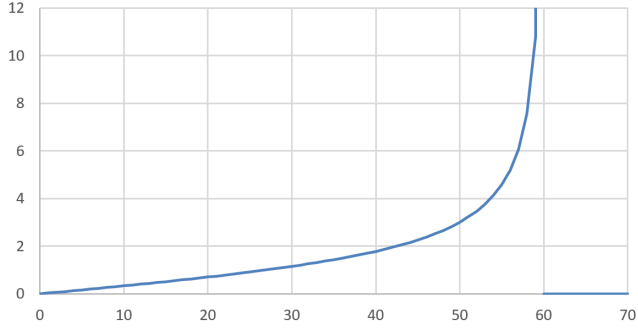

Figure 1.  $h_{60}(x)$

We can see that there is a sharp increase in the probability when the interval length approaches the diameter of the discs. So instead of taking the longest interval, we can take the most probable interval as the size of the discs. This idea immediately leads to an algorithm for object size determination: calculate the histogram of segment lengths in horizontal direction (any other direction will do as well) and take the diameter estimate equal to the segment length of the maximum histogram value. Rounding errors obtained from observing discs drawn on a pixel grid do not influence the general shape of the distribution function but its maximum is about 2 pixels to the left of the diameter of the disc. That is caused by the fact that the function $h_d(x)$ goes to infinity when $x$ approaches $d$ what would be impossible in the discrete case. What happens is that intervals of length approaching $d$ are rounded down leading to the mentioned shift. We will deal with this phenomenon later, together with discs of varying size.

Such algorithm also work in the case of overlapping discs. Overlaps join some segments introducing mostly long ones and if the overlap is not too large, the maximum of the histogram remains the same. To see how much overlap can be tolerated to be able to reliably extract the correct object size, we made an experiment by generating a progressive number of randomly placed discs of diameter 100 in an image of size 1000×1000 and calculated the histogram of the lengths of white horizontal intervals. We determined the object size as discussed above and reported an error when the diameter estimate was off by more than 3 pixels. We did 100 experiments for each density. The graph in Fig. 2 shows the number of errors for each density. Examples of the images used are shown in Fig. 3. An example of interval histogram is shown in Fig. 4.

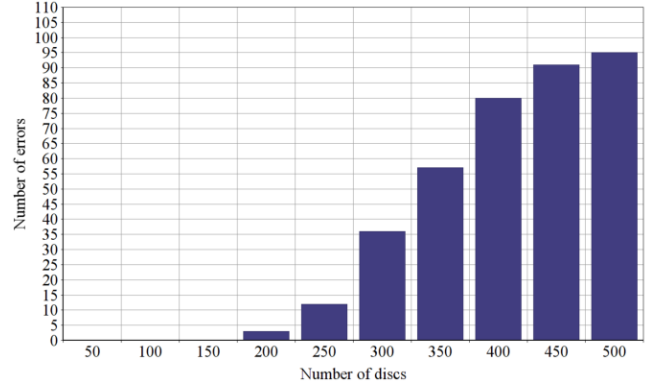

Figure 2.  Number of errors for each density

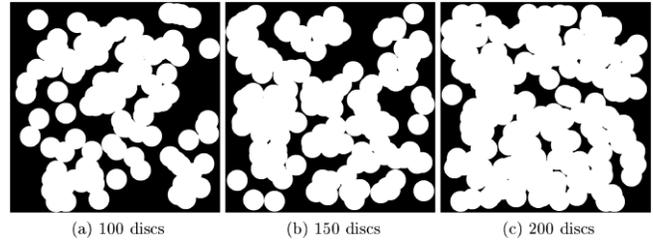

(a) 100 discs    (b) 150 discs    (c) 200 discs

Figure 3.  Randomly places discs

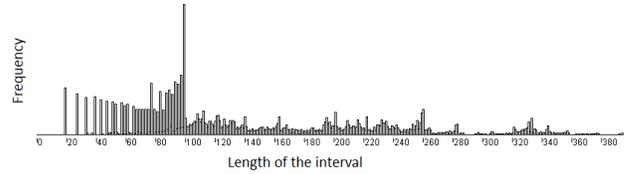

Figure 4.  Number of intervals with each length in the image with 150 discs

This experiment shows that the size can be extracted without any error if we place up to 150 discs in the image. Such case corresponds to a fairly crowded image where objects occupy about 65% of the image and the object overlap is about 45%.

Now let us consider a case when object sizes are not ideally circular and their sizes can differ a bit. Let us take an example image containing 20 circular objects with diameters ranging from 61 to 80. A function specifying its segment length distribution is presented in (2); it is shown in Fig. 5(a). In real examples the distribution is much more jagged, like in Fig. 4, therefore we apply a mean filter of length 11 (this constant is chosen experimentally). The function of our model becomes (3) and is shown in Fig. 5(b).

$$g_{61,80}(x) = \sum_{d=61}^{80} h_d(x) \qquad (2)$$

$$g'_{61,80}(x) = \frac{\sum_{i=x-5}^{x+5} g_{61,80}(i)}{11} \qquad (3)$$

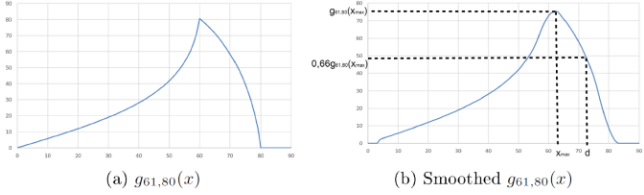

Figure 5. The distribution function for differently sized

The average diameter of the circles in this example is 70.5, but the maximum of g' is located at x=63. Therefore, we must introduce a correction factor. After doing so and rewriting formula for a general case the formula for diameter calculation becomes (4).

$$d = min\{x | x > x_{max} \land g'(x) \leq \alpha g'(x_{max})\}, \quad (4)$$

where $g'(x)$ is the smoothed segment length distribution function, $x_{max}$ is the $x$ value of its maximum and $\alpha$ is a correction coefficient that depends on the distribution of the object sizes and can be found experimentally for every data set. We found that $\alpha \in [0.66, 0.8]$ works well for the test cases described below.

III. THE ALGORITHM

Now we are ready to state the whole algorithm for object counting, it consists of the following steps:

1. Image segmentation – each pixel is classified either belonging to background (black) or an object (white).
2. Image enhancement – a median filter is applied to reduce split segments. We used 5×5 pixel mask.
3. Calculation of object sizes.
(a) Create a histogram of the lengths of the horizontal segments.
(b) Smooth the histogram with a mean filter. We use a mask of size 11.
(c) Find the maximum of the histogram and calculate the estimated diameter by (4).
4. Calculation of the number of objects.

First two steps are generally accepted in object counting [10]. Depending on the quality and characteristics of the images, different methods can be used in image enhancement and segmentation steps. In our tests we have used two popular segmentation methods – Otsu's segmentation and Region growing method [9, 3].

In step 4 we assume that objects can be approximated by circles and their count is calculated by dividing the number of white pixels $S$ in the image by the area of a single object obtained from its diameter $d$, see (5).

$$c = \frac{S}{\pi(\frac{d}{2})^2}, \quad (5)$$

In case of significantly overlapping objects this formula may not be accurate but it still can be used if we adjust the coefficient $\alpha$, which is selected experimentally anyway.

IV. EVALUATION

We evaluated our algorithm on 4 different sets of images – human bone cells with highlighted nuclei under microscope, oranges put on a black cloth, elliptic pills put on a wooden board and round pills put closely together on a black background. Example images are presented in Fig. 6. Images with nuclei were taken from [1]. Other three sets were created by the authors. Every set that we used consisted of 50 different images with a different object placement in every image. We manually counted the number of objects in every image and then compared the count with the number of objects obtained by the proposed algorithm. For tests with nuclei we used $\alpha$=0.66, for all other tests – $\alpha$=0.8. All other parameters were left at their default values. The average error on these data sets was: nuclei – 7.9%, oranges – 1.6%, elliptic pills – 6.6%, round pills – 3.9%.

To compare the results, we also implemented object counting using Watershed method that might be the most appropriate of the known methods for the given image types (images that contain circular or elliptical, partly overlapping objects with partially varying sizes, shapes and orientations). The average error on the same data sets was: nuclei – 6.3%, oranges – 0.8%, elliptic pills – 5.4%, round pills – 1.3%. Results of both methods are very close, especially considering the ease of implementation and linear time complexity of our method.

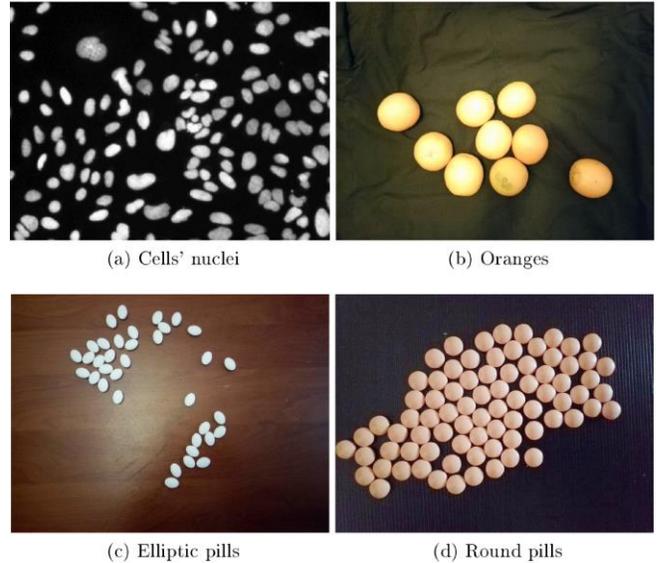

Figure 6. Examples of images used for testing

V. CONCLUSION

A new method for statistical object counting from images was presented in this paper. The method can count circular or elliptic objects with similar sizes. Unlike most of the known methods, the new method is not identifying every object, it uses statistical data obtained from an image instead. The method was implemented and tested on 4 different sets of images. The error on all sets was lower than 8% that can be enough for many practical applications. The presented method has several advantages:

1. It is able to count objects in images where many objects are touching or are partially overlapping.

2. The method does not require complicated tuning to each kind of image or object type. The only parameters that were changed for different datasets were the segmentation method and the parameter $α$. All other parameters remained at their default values. Even though objects in different sets are of different size and shape, the presented method could count them accurately. Parameter $α=0.66$ was used only for one dataset. If we use value $α=0.8$ also for this dataset, we get error of 16.3% that still can be acceptable.

3. The method is simple and fast, it is performing in linear time from the number of pixels in the image.

The method assumes that objects are placed more or less randomly. If specific patterns of touching objects are frequent, a false maximum of segment length distribution may emerge and the method can fail to produce accurate results. We suppose that this obstacle can often be overcome by using segments from different directions (not only horizontal). Another limitation is that the objects should not contain large holes. In case of holes larger than the size of the chosen median filter in step 2, many segments can be split and a wrong object size can be chosen.

ACKNOWLEDGMENT

This work was supported by Latvian State Research programme NexIT project No.2.

REFERENCES

[1] Cell image library, http://www.cellimagelibrary.org/images/45778

[2] Costa, C.M., Yang, S.: Counting pollen grains using readily available, free image processing and analysis software. Annals of Botany (2009)

[3] Gonzalez, R., Woods, R.: Digital Image Processing. International Edition, Prentice Hall, New Jersey (2002)

[4] Guo, X., Yu, F.: A method of automatic cell counting based on microscopic image. In: Intelligent Human-Machine Systems and Cybernetics (IHMSC), 2013 5th International Conference on. vol. 1, pp. 293-296 (2013)

[5] Jambhekar, N.D.: Red blood cells classification using image processing. Science Research Reporter 1(3), 151-154 (2011)

[6] Jernot, J.P.: Thése de Doctorat és Science. Ph.D. thesis, Université de Caen, France (1982)

[7] Mao-jun, S., Zhao-bin, W., Hong-juan, Z., Yi-de, M.: A new method for blood cell image segmentation and counting based on pcnn and autowave. In: Communications, Control and Signal Processing, 2008. ISCCSP 2008. 3rd International Symposium on. pp. 6-9 (2008)

[8] Mogra, M., Bansel, A., Srivastava, V.: Comparative analysis of extraction and detection of rbcs and wbcs using hough transform and k-means clustering algorithm. International Journal of Engineering Research and General Science 2(5), 670-674 (2014)

[9] Otsu, N.: A Threshold Selection Method from Gray-level Histograms. IEEE Transactions on Systems, Man and Cybernetics 9(1), 62-66 (1979)

[10] Pandit, A., Rangole, J.: Literature review on object counting using image processing techniques. International Journal of Advanced Research in Electrical, Electronics and Instrumentation Engineering 3(4), 8509-8512 (2014)

[11] Russ, J.C.: The image processing handbook. CRC press, London, New York (2011)

[12] Tangsuksant, W., Pintavirooj, C., Taertulakarn, S., Daochai, S.: Development algorithm to count blood cells in urine sediment using ann and hough transform. In: Biomedical Engineering International Conference (BMEiCON), 2013 6th. pp. 1-4 (2013)